\newcommand{\sasha}[1]{}
\newcommand{\joel}[1]{}
\newcommand{\remi}[1]{}
\newcommand{\jzl}[1]{}
\newcommand{\sko}[1]{}
\newcommand{\skot}[1]{}
\newcommand{\remove}[1]{}
\newcommand{\ywu}[1]{}

\newcommand{\parshrinky}{}
\newcommand{\secshrinky}{}
\newcommand{\subsecshrinky}{}
\newcommand{\figshrinky}{}

\documentclass{article}

\usepackage{microtype}
\usepackage{graphicx}
\usepackage{booktabs} 
\usepackage{xcolor}
\usepackage{caption}
\usepackage{subcaption}
\usepackage{amsmath}
\usepackage{amsfonts}

\usepackage{hyperref}

\usepackage[accepted]{icml2020}

\icmltitlerunning{Options as responses}

\begin{document}

\twocolumn[
\icmltitle{OPtions as REsponses: \\ Grounding Behavioural Hierarchies in Multi-Agent Reinforcement Learning}



\icmlsetsymbol{equal}{*}

\begin{icmlauthorlist}
\icmlauthor{Alexander Sasha Vezhnevets}{equal,dm}
\icmlauthor{Yuhuai Tony Wu}{equal,dm,goo}
\icmlauthor{Maria Eckstein}{ed,dm}
\icmlauthor{R\'emi Leblond}{dm}
\icmlauthor{Joel Z. Leibo}{dm}
\end{icmlauthorlist}

\icmlaffiliation{dm}{DeepMind, London, UK}
\icmlaffiliation{goo}{University of Toronto, Canada}
\icmlaffiliation{ed}{University of California Berkeley, USA}

\icmlcorrespondingauthor{Alexander Sasha Vezhnevets}{vezhnick@google.com}

\icmlkeywords{Machine Learning, ICML}

\vskip 0.3in
]



\printAffiliationsAndNotice{\icmlEqualContribution} 

\begin{abstract}
This paper investigates generalisation in multi-agent games, where the generality of the agent can be evaluated by playing against opponents it hasn't seen during training.
We propose two new games with concealed information and complex, non-transitive reward structure (think rock/paper/scissors).
It turns out that most current deep reinforcement learning methods fail to efficiently explore the strategy space, thus learning policies that generalise poorly to unseen opponents.
We then propose a novel hierarchical agent architecture, where the hierarchy is grounded in the game-theoretic structure of the game -- the top level chooses strategic responses to opponents, while the low level implements them into policy over primitive actions.
This grounding facilitates credit assignment across the levels of hierarchy. 
Our experiments show that the proposed hierarchical agent is capable of generalisation to unseen opponents, while conventional baselines fail to generalise whatsoever.
\end{abstract}

\section{Introduction}
\label{sec:intro}
\secshrinky

The outstanding feature of human intelligence is its generality. 
Although great strides in deep RL have been made, agents are often not capable of generalisation beyond trivial changes in the environment~\cite{leike2017ai, machado2018revisiting,cobbe2019quantifying}, which limits the practical impact.
Consequently, researchers have begun advocating borrowing experimental design from supervised learning with explicitly separated testing and training version of the environment~\cite{machado2018revisiting, cobbe2019quantifying, racaniere2017imagination}.
In this paper we want to address the problem of building agents with \textit{general competence} over a wide varieties of tasks, including those not experienced during training.
It is an open question how these tasks could be created for either training or evaluation.
Here we pursue an avenue for investigating generalisation and problem generation, which comes from multi-agent research. 
As argued in~\cite{leibo2019autocurricula}, multi-agent systems can display intrinsic dynamics arising from competition and cooperation that provide a naturally emergent and dynamically growing set of problems.
A solution of one multi-agent task begets a new task, continually generating novel challenges. 
Recent success stories like AlphaGo~\cite{silver2016mastering} and AlphaStar~\cite{vinyals2019grandmaster} have demonstrated generalisation by playing directly against professional human players, which they had no experience playing before.
AlphaStar~\cite{vinyals2019grandmaster} has explicitly generated an ever growing set of opponents to both train and evaluate the agent against.
Although StarCraft is a great challenge, its computational demands and game complexity can be prohibitive for fundamental research.

The first contribution of this paper are two grid world multi-agent games with simple implementation, yet complex multi-agent dynamics with non-transitive reward (think rock/paper/scissors) and concealed information.
We can empirically measure generalisation by playing policies learnt by different agents against each-other and a pre-trained hold-out set of policies.
If the agent has understood the rules of the game and explored the space of strategies well, its policy should fare well against a novel adversary.
Our experiments show that most common deep RL agent architectures~\cite{espeholt2018impala,foerster2018counterfactual,lowe2017multi} generalise poorly, even if designed for multi-agent specifically~\cite{foerster2018counterfactual,lowe2017multi}.

Our second contribution is a novel hierarchical agent architecture, which is grounded in the game-theoretic structure of the game.
To discover that structure, we factorise the value function over a latent representation of the concealed information and then re-use this latent space to factorise the policy. 
Low-level policies (options) are trained to respond to particular states of other agents grouped by the latent representation, while the top level (meta-policy) learns to infer the latent representation from its own observation and thereby select the right option.
This architecture, coined OPtions as REsponses (OPRE), generalises better by separately extracting strategic and tactical knowledge of the game into separate levels of hierarchy.
It then generates new behaviours by recombining learnt options given an unfamiliar opponent.  
Experiments show that OPRE significantly outperforms baselines, when evaluated against them directly or when measured against held-out opponents that the agents did not see in training.

The rest of the paper is organised as follows. 
Section~\ref{sec:domains} introduces multi-agent games with concealed information and non-transitive reward. Section~\ref{sec:hrl4marl} formulates the hierarchical approach to solve those domains. 
We introduce the OPRE model in Section~\ref{sec:hrl4marl}, followed by a review of the related work in Section~\ref{sec:related_work}.
Experimental results are presented in Section~\ref{sec:experiments}. 
We conclude in Section~\ref{sec:conclusions}.

\section{Multi-Agent RL games with concealed and partial information}
\label{sec:domains}

Consider a modified game of rock-paper-scissors, where each strategy (e.g. rock), rather than being executed in one shot as an atomic action, requires a long sequence of actions that, taken as a whole, \emph{constitutes} the chosen strategy.
In keeping with the terminology introduced in \cite{leibo2019autocurricula}, we call such a policy an \emph{implementation} of the strategy.
Both strategic decisions and their implementations must be learned by reinforcement learning.

We propose two stateful extensions of the classic matrix game rock-paper-scissors with increasing complexity, called Running With Scissors (RWS) (Section~\ref{sec:RWS}) and RPS Arena (Section~\ref{sec:RPS_Arena}).
In these \emph{spatialised} extensions, players move around on a 2D grid, only observing a small window around their current location (the environment is \emph{partially observed}).
They can pick up resources (i.e. rocks, papers and scissors), which are scattered semi-randomly, and can confront their opponent(s).
Once the confrontation occurs (via a tagging mechanic) then the payoffs accruing to each player are calculated according to the standard anti-symmetric matrix formulation of rock-paper-scissors (see, e.g., \citet{hofbauer2003evolutionary}).
In these games, to implement a strategy---say the `rock' pure strategy---an agent must seek out tiles with rock resources, while avoiding collecting non-rock resources, and then confront an opponent.

RWS is a two-player game with a single reward event at the end of the episode. 
RPS Arena is a five-player game with much longer episodes, more complex dynamics and multiple reward events. 
Both domains are partially observable, spatially and temporally extended multi-agent games with a game-theoretic payoff structure.
Contrary to many other multi-agent studies~\cite{foerster2018counterfactual,lowe2017multi,ahilan2019feudal,hughes2018inequity}, the tasks are not cooperative (agents do not share a reward signal) and do not feature a communication channel.

All the rich game theoretic structure of the rock-paper-scissors game is maintained in RWS and RPS Arena. 
In addition, agents also have to cope with the complexity of learning implementations of strategies. 
For example, the game's intransitive response dynamics (rock beats scissors, which beat paper, which in turn beats rock), may induce endlessly recurring cyclic reinforcement learning challenges~\cite{balduzzi2018re, leibo2019autocurricula, omidshafiei2019alpha}. 
Furthermore, because they have imperfect information, agents cannot easily figure out their opponent's strategy and select the appropriate best response.
However, they have ways to gather information about their opponent's intent.
In Running With Scissors, agents are able to move around the map and observe what resources are missing, implying they were already collected by the adversary. 
A reasonable way to play this game is to first attempt to scout around and determine to what resource the adversary has committed itself, and to then implement the best response to that strategy. 
The crucial point in this setup is that these strategies (sub-behaviours) are all naturally grounded -- the top level deals with the choice of strategy as a response to the opponent's strategy, and the bottom level with its implementation. 
They are also precisely what one hopes would be discovered by any option-discovery hierarchical reinforcement learning (HRL) algorithm.

\subsection{Running with scissors (RWS)}
\subsecshrinky
\label{sec:RWS}

\begin{figure}[h]
\centering

{\includegraphics[width=\columnwidth]{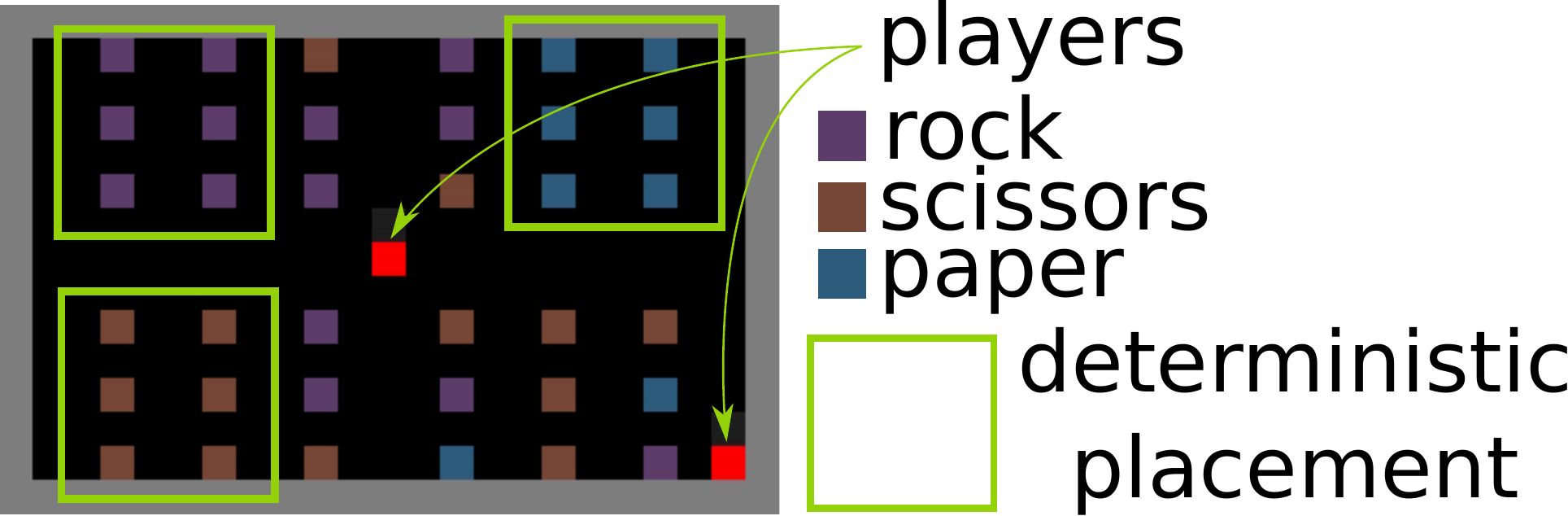}}
  \caption{ Running With Scissors (RWS) illustrated. This is the initial configuration of the grid. Players spawn at any random location, except for those occupied by resources. Some resources are placed deterministically (shown by green rectangles), others spawn randomly. \label{fig:RWS_ill} }
\end{figure}

Figure~\ref{fig:RWS_ill} visualises the initial grid configuration of Running with Scissors.

Formally, RWS is a Markov game \cite{littman1994markov} -- Markov games are to matrix games as Markov decision processes are to bandit problems.
The grid has dimensions $13\times21$; each of the two players starts at a random position on the grid and can observe a $4\times4$ area around itself.
There are $36$ evenly spaced resources (i.e. rock, scissors or paper) tiles on the grid, $18$ of which always have the same resource while others get a random one.
When a player steps on a tile with the resource, it picks it up -- the resource is removed from the grid and is allocated into the player's inventory $v$, which is initialised to contain one of each resource type.

There are two termination conditions for the episode: i) timeout of 500 steps or ii) one agent tags another. 
Tagging is a special action that highlights a $3\times3$ area in front of the tagging player and is successful if the other player is in that area.
Which agent was tagged does not matter; the outcome is the termination of the episode and scoring.
The rewards accruing to each player are calculated as follows:
$r^0=\frac{v^0}{||v^0||}M\left(\frac{v^1}{||v^1||}\right)^T=-r^1$, where $v^0$ and $v^1$ are the respective inventories and $M$ is the anti-symmetric matrix of the game $M=100\cdot $\begin{tiny}$\begin{bmatrix}
    0 & -1 & 1  \\
    1 & 0 & -1 \\
    -1 & 1 & 0 
\end{bmatrix}$ \end{tiny}.

To get a high reward an agent should correctly identify what resource its opponent is collecting (e.g. rock) and collect a counter (e.g. paper). 
In addition, the rules of the game are not assumed given; the agents must explore to discover them. 
Thus it is simultaneously a game of \emph{imperfect} information---each player possesses some private information not known to their adversary (as in e.g. poker \cite{sandholm2015solving})---and \emph{incomplete} information---lacking common knowledge of the rules \cite{harsanyi1967games}.

\subsection{RPS Arena}
\label{sec:RPS_Arena}
\subsecshrinky

\begin{figure}[h]
{\includegraphics[width=\columnwidth]{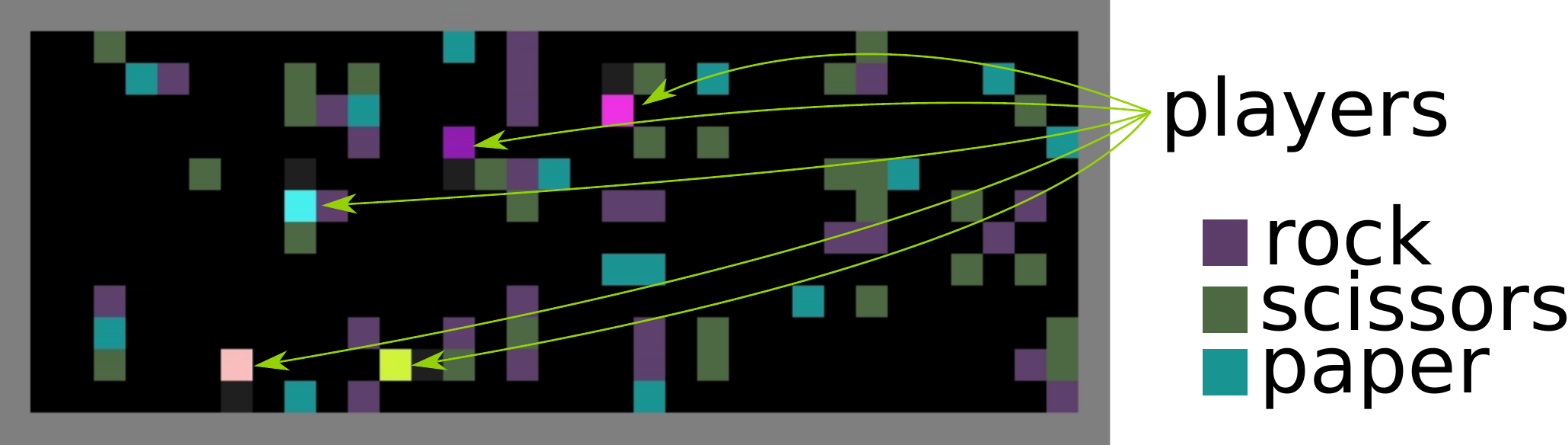}}
  \caption{ RPS Arena illustrated. Resources are placed randomly and re-spawn over time. There are 5 players in total. \label{fig:rwa}. }
\end{figure}

We propose a more challenging version of RWS, coined RPS Arena, that extends it in the following ways: i) the world is larger $13\times42$ \remi{(check this) and remove comment} ii) there are 5 players iii) resources (R/P/S) are scattered randomly and regenerate. 
The reward is collected by agents tagging each other --- their inventories reset, the reward is calculated as in RWS but the episode does not terminate; additionally, the losing agent is randomly teleported and frozen for 50 steps.
It is a lot harder for an agent to predict the inventory of the opponent, since there are 4 of them now and the map is larger.

\subsection{Pre-training hold-out competitors.}
\label{subsec:holdout}
To test generalisation we need a held-out set of pre-trained competitors, which is generated as follows. 
We trained a set of 14 agents in simultaneous population-play.
12 of them were augmented with pseudo-rewards during training that incentivised them to learn a particular pure strategy (4 per strategy).
Incentivised agents receive a pseudoreward of $+10$ for collecting each resource of a specific type and $-5$ for collecting any other resource.
The remaining two out of the 14 simultaneously learning agents were given no pseudorewards for robustness.
This was done for both domains.

\section{Hierarchical RL for MARL}
\label{sec:hrl4marl}
The key issue in HRL is credit assignment for high-level choices vs their execution on the low level --- if the reward is low, is this due to a wrong choice of strategy or its faulty execution? 
One way to address that is to ground high-level decisions in something exogenous to the agent, so that the success or failure of the high-level policy can be inferred separately from the success or failure of the implementation.
For example, Feudal RL~\cite{dayan1993feudal, vezhnevets2017feudal} grounds high-level decisions in the state of the environment, so the choice of goal state and whether it was achieved are verifiable independently.
Here, for the multi-agent case we propose to ground the high level in the concealed state of the opponent(s)---the high level estimates what that state is and the low level implements a response to it (e.g. if my opponent is rock, play paper).

To cast the multi-agent learning problem as hierarchical RL, we build a latent representation, $z$, of the concealed information.
The proposed hierarchical agent, OPRE (OPtions as REsponses), learns a best response policy to a particular configuration of $z$ at the low level of the hierarchy, while the top level of the hierarchy estimates $z$ and thereby picks an option.
We learn the latent representation $z$ by factorising the value function over the concealed information using a mixture model. 
For training, we assume that we have access to this concealed information \textit{in hindsight} for the purpose of generating learning signal;\footnote{This is a reasonable assumption, akin to a sportsman watching a replay of the game to analyse their play a posteriori. StarCraft 2 players routinely watch replays with opponent play revealed during championships.} we \textit{never} use concealed information to generate any behaviour.
This is similar to hindsight policy gradient~\cite{rauber2017hindsight} and experience replay ~\cite{andrychowicz2017hindsight} and other counterfactual RL works~\cite{buesing2019woulda,foerster2018counterfactual,lowe2017multi}.
This grounding helps the agent distinguish between two types of mistakes: i) wrongly guessing the concealed information, thereby picking an inadequate strategy (e.g. rock vs paper) or ii) picking the right strategy, but executing it poorly. 

\subsection{Formal definition of MARL}
\label{sec:setup}
\secshrinky

We consider an $N$-player partially observable Markov game $\mathcal{M}$ defined on a finite set of states $\mathcal{S}$.
The observation function $\mathcal{X} : \mathcal{S} \times \{1, . . . , N\} \rightarrow \mathbb{R}^d$, specifies each player's $d$-dimensional view on the state space.
In each state, each player $i$ is allowed to take an action from its own set $\mathcal{A}^i$.
Following their joint action 
$(a^1, . . . , a^N) \in \mathcal{A}^1 \times \! . . . \! \times \mathcal{A}^N$,
the state obeys
the stochastic transition function 
$\mathcal{T} : \mathcal{S} \times \mathcal{A}^1 \times \! . . . \! \times \mathcal{A}^N \rightarrow \Delta(\mathcal{S})$
($\Delta(\mathcal{S})$ denotes the set of discrete probability distributions over $\mathcal{S}$) and every player receives an individual reward defined as 
$r^i: \mathcal{S} \times \mathcal{A}^1 \times . . .  \times \mathcal{A}^N \rightarrow \mathbb{R}$
for player $i$. 
Finally, let 
$\mathcal{X}^i = \{\mathcal{X}(s, i)\}_{s \in \mathcal{S}}$ 
be the observation space of player $i$.

Each agent learns independently through their own experience of the environment by optimising for their own individual return, without direct communication to other agents.
Notice that this is different from a large part of the recent work on multi-agent RL~\cite{foerster2018counterfactual,rashid2018qmix,iqbal2018maac}, where agents are working towards a common goal by optimising a shared reward (with the notable exception of~\cite{lowe2017multi}, which allows for cooperative-competitive environments), often coordinating via an explicit communication channel.

The rest of the paper considers an ``egocentric'' view of one of the agents in the environment, thereby we will denote its observations and actions simply as $x$ and $a$. 
The information concealed from the agent is denoted as $x'$. 
Here, we define it as the observations of all other agents. 
So for the agent of interest, $x'=\bigcup_{x^j \neq x} x^j$.

\subsection{OPRE model}
\label{sec:model}
\secshrinky

\begin{figure}[th]
  \centering
{\includegraphics[width=\columnwidth]{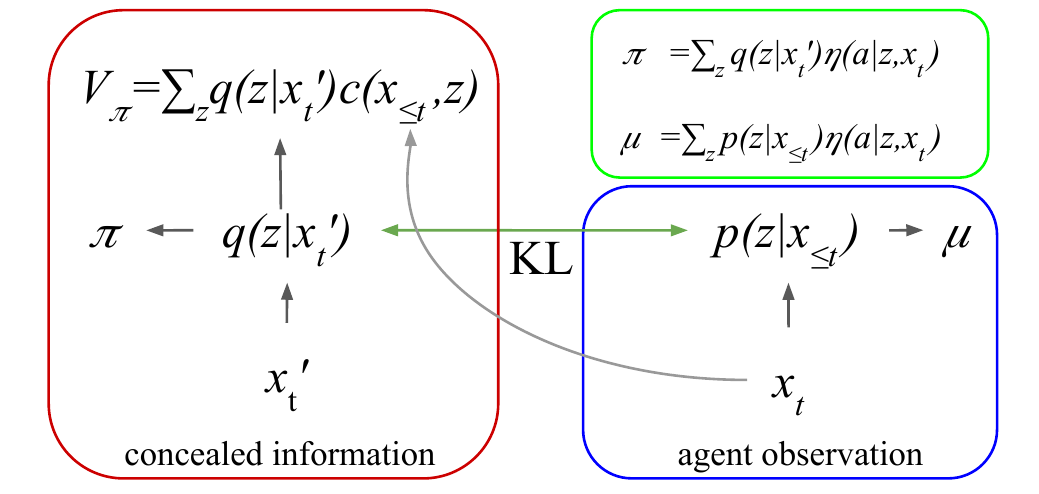}}
  \caption{ OPRE illustration. The first module (red box) uses concealed information, available in hindsight, to factorise the value function and target policy. The second module (blue box) produces behaviour policy, by approximating the latent distribution $q$ \emph{without} using the concealed information. \figshrinky \label{fig:model_schema}}
\end{figure}

This section introduces the OPtions as REsponses (OPRE) model -- a hierarchical agent for multi-agent RL. 
Figure~\ref{fig:model_schema} illustrates the structure of the model which consists of two parts. 
The first part (Figure~\ref{fig:model_schema}, red box on the left) builds the latent space by factorising the value function over the concealed information, while the second part (Figure~\ref{fig:model_schema}, blue box on the right) re-uses that latent space to factorise the agent's policy into options. 
Both value and policy are mixture models, where a categorical latent variable selects the mixture component.
These mixture models are synchronised via a KL term, which pulls the mixture weights to be similar in both for each step.

\parshrinky
\paragraph{Building the latent space via value function factorisation.} We model the value function of the agent as a mixture model, where the mixture weights $q(z|x'_t)$ depend on the concealed information $x'_t$, while the mixture components $c(x_{\leq t},z)$ depend on the agent's own history of observations $x_{\leq t}$:
\begin{equation}
    V_{\pi}(x_{\leq t},x'_t)=\sum_{z}q(z|x'_t)c(x_{\leq t},z).
    \label{eq:value_fact}
\end{equation}
Here the latent variable $z$ is assumed to be categorical and can be marginalised away.
This construction compresses the concealed information into the distribution $q(z|x'_t)$ over the latent space $z$, but only focusing on the part which is \textit{relevant to representing the value}. 
\sko{Maybe try to rephrase previous for increased clarity/precision.}

\parshrinky
\paragraph{Inducing options.} 
The key idea behind OPRE is to re-use the same latent space to factorise the policy. 
Consider first this mixture policy:
\begin{equation}
    \pi(a|x_{\leq t},x'_t)=\sum_{z}q(z|x'_t)\eta(a|x_{\leq t},z),
    \label{eq:target_policy}
\end{equation}
where $\eta(a | x_{\leq t},z)$ is a mixture component of the policy -- an option.
In this form, an option is a behavioural response to a particular latent representation $z$ of concealed information $x'$.
$z$ corresponds to a group of opponent states for which a certain mixture component is activated (i.e. for which $q(z|x'_t)$ is high).
To compute $\pi(a | x_{\leq t},x'_t)$ the agent needs to know $q(z|x'_t)$, which depends on the concealed information $x'_t$. 
However, by definition, this quantity is not available to the agent.
Therefore, we introduce the behaviour policy, which approximates it from the agent observation alone:
\begin{equation}
    \mu(a|x_{\leq t})=\sum_{z}p(z|x_{\leq t})\eta(a|x_{\leq t},z).
    \label{eq:behaviour_policy}
\end{equation}
In this way, $p(z|x_{\leq t})$ becomes the policy over options, which is learned separately as described below.

\subsection{Learning}
\label{sec:learning}
\subsecshrinky
This section describes how to train OPRE agents. 
We use an IMPALA~\cite{espeholt2018impala} based computational setup for training all of our agents.
A few hundred CPU actors generate game trajectories, which are then batched and consumed by a GPU learner (one per unique agent).
The learner receives temporally truncated sequences of $100$ steps of trajectories in batches of $16$.
The parameters are periodically synced between the actors and the learner.
The concealed information about other agents is only available to the learner (the actors, which generate behaviour, do not use concealed information). 
V-trace is used to correct for off-policy due to the lag of parameter synchronisation on actors and learners.
Algorithm~\ref{OPRE_psuedocode_actor} and~\ref{OPRE_psuedocode_learner} present pseudo-code for actor and learner respectively.

\begin{algorithm}[tb]
    \caption{Training OPRE: Actor}
    \label{OPRE_psuedocode_actor}
\begin{algorithmic}[1]
\small{
\STATE {\bfseries Input:} ($N,K$) // takes max number of steps and number of agents
\WHILE{$n < N$} 
\IF {$\textit{time to sync}$ \textbf{or} $n=0$}
\STATE $\Theta = \{\theta_k \}^K_{k=1} \gets \Theta_{\textit{step}}$ // sync parameters with learner
\ENDIF
\STATE $i \gets \textit{random}(K)$, $j \gets \textit{random}(K)$ // draw two agents to play at random
\STATE $\tau \gets \textit{Play}(\mu_{\theta_i}, \mu_{\theta_j})$ // generate trajectory by playing behaviour policies  
\STATE $\textit{Queue}(\tau)$ // queue the trajectory for the learner
\STATE $n \gets n_{learner}$ // sync learning step counter with the learner
\ENDWHILE
}
\end{algorithmic}
\end{algorithm}

\begin{algorithm}[tb]
    \caption{Training OPRE: Learner}
    \label{OPRE_psuedocode_learner}
\begin{algorithmic}[1]
\small{
\STATE {\bfseries Input:} ($N, k, T, B$) // takes max number of steps, agent index, batch dimensions
\STATE $\theta \gets \textit{random}()$
\STATE $n \gets 0$
\WHILE{$n < N$} 
\STATE $\beta \gets \textit{Dequeue}(k, B, T)$ // get a batch for a respective agent of $B\times T$ size 
\STATE $\nabla \pi_{\theta}, \nabla V_{\theta} \gets \textit{V-trace}(\beta,\theta)$ // get target policy $\pi$ gradients with V-trace \cite{espeholt2018impala}
\STATE $\nabla \textit{Reg} \gets \nabla(H^T(q_{\theta}) - H^B(q_{\theta}) - H(\pi_{\theta})) $ // gradients from regularisation
\STATE $\theta \gets \textit{BackProp}(\nabla \pi_{\theta} + \nabla V_{\theta} + \nabla\textit{KL}(q_{\theta}||p_{\theta}) + \nabla \textit{Reg})$ // update parameters 
\STATE $n \gets n + T \cdot B$ // increment step
\ENDWHILE
}
\end{algorithmic}
\end{algorithm}

\parshrinky
\paragraph{Learning the policy over options} $p(z|x_{\leq t})$ is done via minimising $KL(q||p)$. 
The agent tries to learn the model of its opponents' behaviour by approximating $q(z|x'_t)$ via $p(z|x_{\leq t})$. 
This makes learning the high-level policy straightforward, since there is a clearly defined target $q$ for the policy over options $p$.

\parshrinky
\paragraph{Credit assignment learning options}
Before we describe how the set of individual option policies $\eta$ is learnt, we first note how the multi-agent aspect complicates credit assignment.
The outcomes the agent is learning from are the product of complex interactions between its own states and actions and the states and actions of others.
The same actions by an agent could lead to both high and low reward, depending on the opponents' actions. 
For example, if agent chooses to play a `rock' strategy against an opponent who plays `paper', rather than correcting its `rock' option to be more like `scissors', we only want it to change its decision to play `rock' in the first place. 
Flat agents are not capable of separating these two signals and are likely to cycle between different strategies, as discussed in~\cite{balduzzi2019open}.

We propose to use the policy gradient of $\pi$ for training options, which is computed off-policy via V-trace~\cite{espeholt2018impala} from the trajectories generated by following the behaviour policy, $\mu$.
This is straightforward, since on the learner we can compute both $\mu$ and $\pi$ by simply marginalising over $z$ via summation ($z$ is categorical).
Importantly, we do not propagate gradients from the policy gradient into $q$, completely leaving its training up to the KL and value function gradients (also learnt via V-trace).
This resolves the credit assignment by separately learning to credit the policy over options $p$ for selecting strategies and the options $\eta$ for executing them. 
Indeed, if $p$ is far from $q$ then the off-policy term from V-trace will dampen the gradient for $\eta$ and the only training signal will be the KL between $p$ and $q$.
Again, we emphasise that we only use the concealed information $x'$ for learning and never use it to run the agent's policy.

This can, potentially, lead to a bias in the policy when $p$ cannot possibly match $q$ -- e.g. the agent's state does not contain enough information to make the right inference about the concealed information. 
As an ablation, we consider a greedier version of OPRE, in which an additional policy gradient from the behavioural policy $\mu$ is propagated into $\eta$; coined \textbf{OPRE mix PG}.
This ablation directly optimises the behaviour policy $\mu$ to maximise reward, while staying close to the target policy $\pi$, which can be thought of as a form of regularisation. 
Peeking ahead, we shall see in the experiments (Section~\ref{sec:experiments}) that this leads to better exploitation of fixed opponents, but worsens generalisation.

\parshrinky
\paragraph{Regularisation.}
We want to encourage options to be smooth in time within an episode (switch less) and diverse across episodes. 
To this end, we use the following regularisation loss: $H^T(q) - H^B(q) - H(\pi)$. 
The first two terms correspond to the maximisation of the marginal entropy of $q$ over a batch and the minimisation of its marginal entropy over time.
The last term maximises the entropy of the target policy to encourage exploration and avoid policy collapse. 
\sko{Perhaps unpack these a little more to increase clarity.}

\section{Related work}
\label{sec:related_work}

Using concealed information counterfactually to improve the critic in multi-agent learning is a technique used in several recent works~\cite{baker2019emergent,vinyals2019grandmaster,foerster2018counterfactual,lowe2017multi}.
In all of these cases it is only used simply as an input to the critic.
To the best of our knowledge, OPRE's hierarchical approach grounded in concealed information is unique; related works either do not assume a hierarchy of policies~\cite{foerster2018counterfactual,lowe2017multi,iqbal2018maac,rashid2018qmix}, function with a centralised manager as top-level~\cite{ahilan2019feudal}, or do not ground their hierarchy in game-theoretic fashion~\cite{tang2018hierarchical}.

Several works use counterfactual knowledge and hindsight to improve credit assignment in single agent setup~\cite{buesing2019woulda, andrychowicz2017hindsight,rauber2017hindsight,harutyunyan2019hindsight}.

We would like to separately note population-based MARL methods like PSRO~\cite{lanctot2017unified,balduzzi2019open}, which build an iterative procedure on top of the basic RL method to explore the strategy space and find an approximate equilibrium. 
These methods are orthogonal to what we present here and could be combined with OPRE; we leave this to future work.

\section{Experiments}
\label{sec:experiments}
\secshrinky

This section presents the experimental evaluation of OPRE and several baselines on the domains described in Section~\ref{sec:domains}.
First, we describe the agents\footnote{We refer to versions of OPRE and the baselines simply as `agents' in the rest of the document.} that we will use as baseline. 
Next, we evaluate how the agents perform when trained directly against a fixed set of pure strategies (Section~\ref{subsec:exp:exploit}).
This setup is, essentially, a multi-task RL problem, where the task (opponent) is not given and has to be inferred. 
Here agents do not have to generalise to unseen opponents, but need to deal with the non-transitivity of the return and with concealed information.
Section~\ref{subsec:exp:generalise} evaluates agents trained in self-play and tested against the held-out pure strategies. 
This tests the agents' generalisation on a clearly defined benchmark, exogenous to all of them. 
In Section~\ref{subsec:exp:tournament} we analyse agents trained in self-play by playing them against each other and investigating the resulting pay-off matrix. 
Apart from simply declaring a winner, we look at the Nash equilibrium and effective diversity~\cite{balduzzi2019open}.

Given the perceptual simplicity of environments, we use a very slim architecture for OPRE. 
The network starts with a 1D convolutional layer with 6 channels, followed by an MLP with $(64,64)$ neurons, then by an LSTM with $128$ hidden neurons.
We construct $16$ policy heads (one for each $\eta$), where each is an MLP with $128$ hidden neurons taking LSTM output as an input.
We use the same convolutional network with another MLP on top to produce $q$ from the opponent observation (concealed information).
The policy over options $p$ is produced via a linear layer from LSTM output.
For RPS Arena, which has several other agents in the game, we process each of their observations with the same convolutional net and then pool using summation. 

\subsection{Baselines} 

Our first baseline uses a standard architecture with a convolutional network base and a LSTM on top, with linear policy and value head (we simply refer to it as \textbf{Baseline}). 
This is a strong, off-the-shelf deep RL agent.
The second baseline is based on~\cite{foerster2018counterfactual}, where a critic has counterfactual access to concealed information; we call it \textbf{Baseline CC}. 
Concealed information is processed in the same way as in OPRE and then concatenated to the value head's input.
This is also similar to~\cite{lowe2017multi}.
Both~\cite{foerster2018counterfactual,lowe2017multi} use concealed information for training, but do not have a hierarchical structure in their agents. 
The hyper-parameters were tuned on the RWS task in the regime of training and evaluating against competitors; we first tuned the parameters of the baseline, then tuned the extra hyper-parameters of OPRE.
The appendix containst all hyper-parameters, their values and precise network parametrisation like layer sizes.
There are several specific ablation of OPRE and extra baselines in Section~\ref{subsec:exp:ablations}, which are designed to validate specific design choices.
We describe them there.

\begin{figure}[h]
\includegraphics[width=\columnwidth]{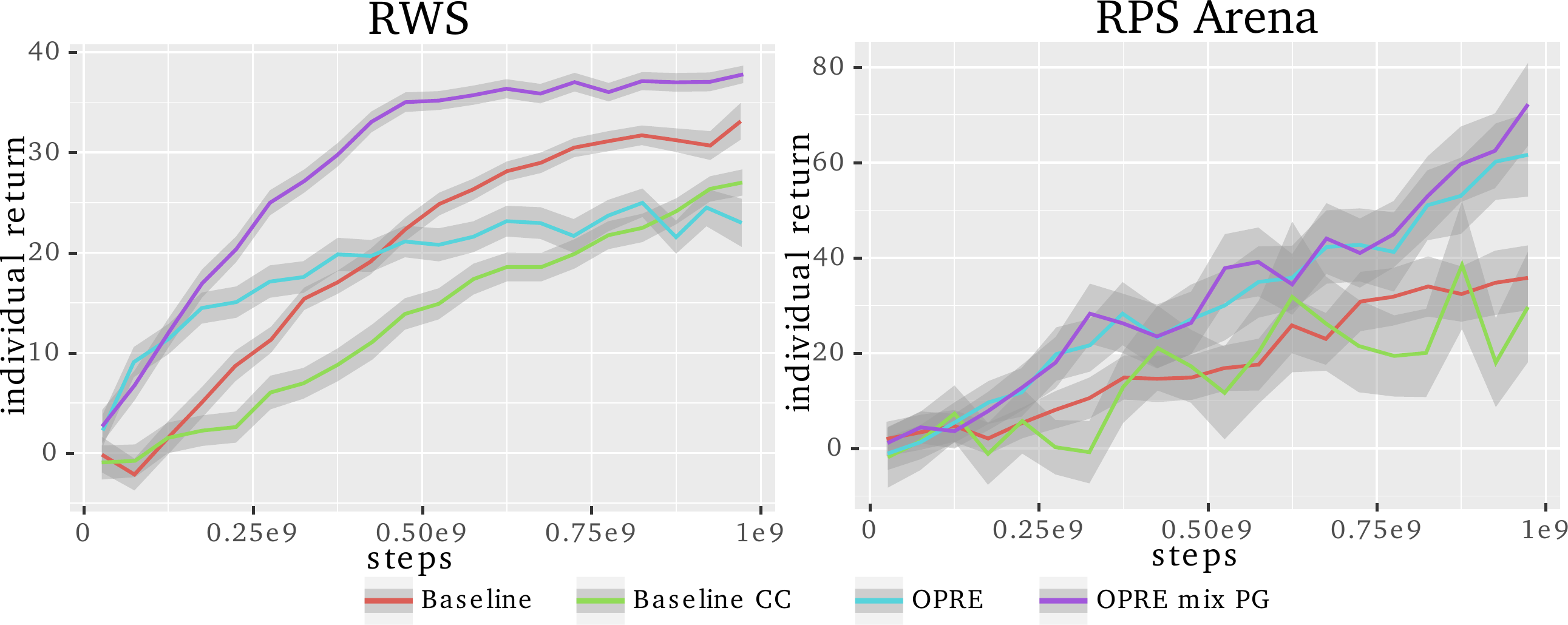}
\caption{ Learning curves for exploiting a fixed set of pure strategies. \textbf{Left:} RWS \textbf{Right:} RPS Arena. \label{fig:exploit_all}}
\end{figure}

\subsection{Exploiting a fixed opponent set}
\label{subsec:exp:exploit}
Here we investigate how good the agents are at simply exploiting a fixed set of opponents (pure strategies from~\ref{subsec:holdout}) on RWS and RPS Arena.
This is a straightforward reinforcement learning problem and we should expect all of the agents to do well.
Figure~\ref{fig:exploit_all} presents the results. 
We report the average reward at a training step with confidence intervals of 0.95, averaged over 6 independent runs.
Indeed, all agents learn to defeat the pure strategies, although differences in data efficiency and final performance are significant.
The clear winner is OPRE mix PG, which is the top agent in both games.
Interestingly, Baseline CC is not performing better than the off-the-shelf Baseline, despite having access to concealed information in the critic.
OPRE performs almost as well as its mix PG version on RPS Arena, but is weaker on RWS.
Notice, how early on OPRE is more data-efficient than the Baseline on RWS, but has worse final performance.
Overal, OPRE agents benefit from using concealed information to guide their policy search, but regular OPRE suffers from bias on the basic version of the domain.

\begin{figure*}[h]
  \centering
{\includegraphics[width=2\columnwidth]{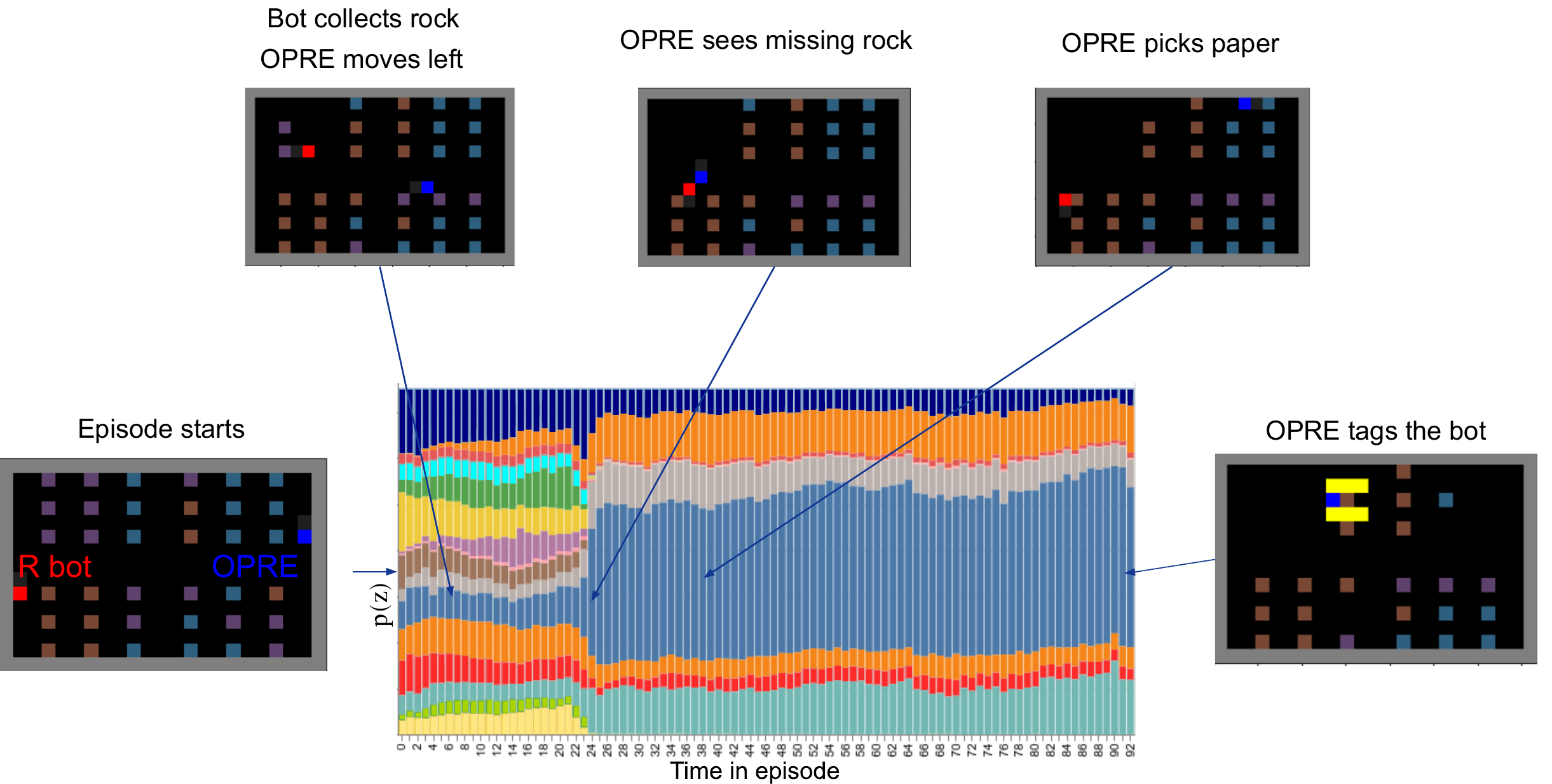}}
  \caption{ The plot presents the evolution of $p(z)$---policy over options---during one episode, where OPRE plays against a pre-trained \textit{rock} opponent (bot). 
  Each colour corresponds an option (element of $z$) and the amount of colour to its weight.
  We render several important moments in the game.
  OPRE and the bot start at opposite sides of the field. 
  Bot proceeds to pick up the rock, while OPRE is moving across the field.
  Once OPRE reaches the opposite side, it looks at rock field and at this moment the $p(z)$ distribution changes drastically, signifying the change in the policy.
  OPRE then proceeds to pick up paper and tag the bot.
  \label{fig:episode_by_frame}}
\end{figure*}

Figure~\ref{fig:episode_by_frame} provides a dissection of one episode against a rock pure strategy, where we visualise evolution of the policy over options $p(z)$. 
This visualisation clearly illustrates how OPRE deliberately investigates the field to gain information on opponent.
The information is then reflected in $p(z)$, which initiates a response behaviour.

\sasha{add conclusions. OPREs are good, since they exploit the structure of the problem.
OPRE is worse of on RWS, because it is biased when p can't match q. 
Baseline CC is surprisingly crap.}

\begin{figure}[h]
  \includegraphics[width=\columnwidth]{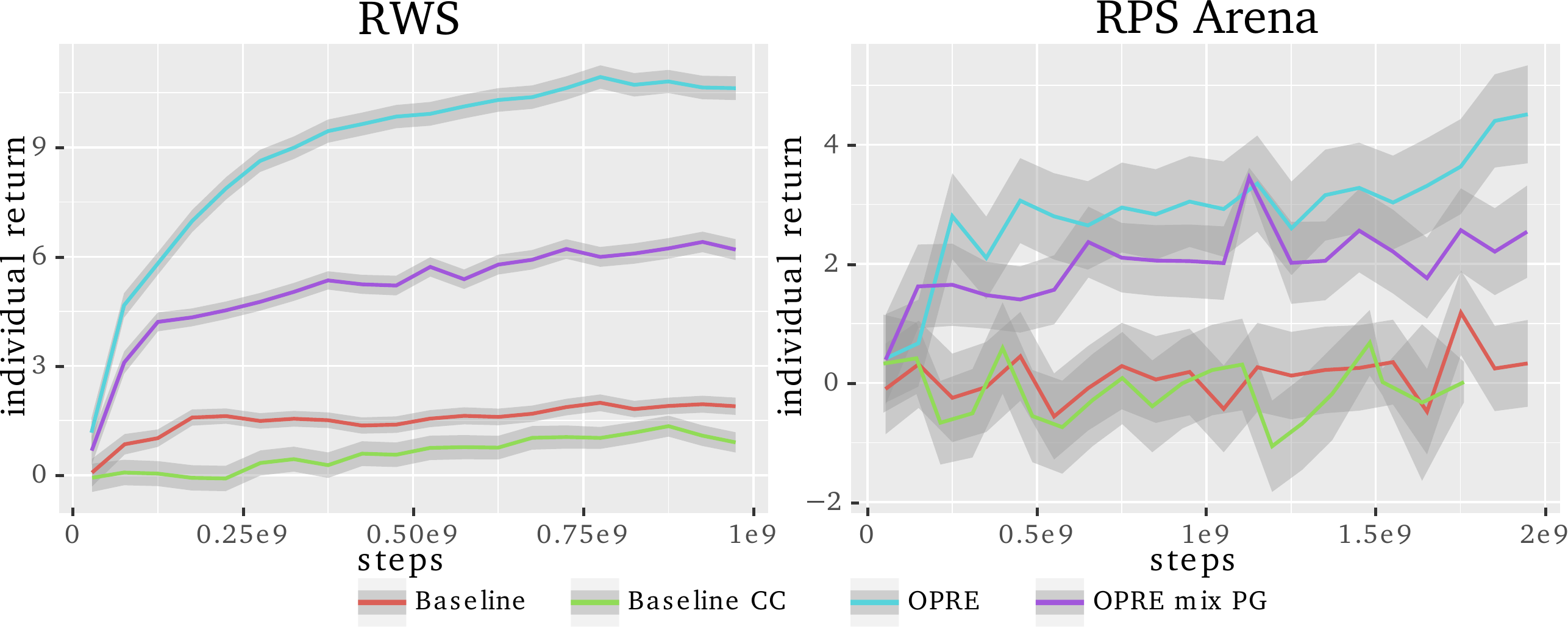}
  \caption{ Generalisation to a hold-out set of pure strategies. \textbf{Left:} RWS; \textbf{Right:} RPS Arena. \label{fig:generalise_all} \figshrinky}
\end{figure}

\subsection{Generalisation to hold-out}
\label{subsec:exp:generalise}

Now we turn to measuring generalisation by testing agents trained in self-play against the held-out pure strategies.
Agents have not encountered the hold-out before and have to generalise.
Here by self-play we mean the following setup: for each agent architecture, we initialise $6$ agents of that architecture using a different random seed; the agents then compete against each other, sampled for a match at random.
All seeds learn simultaneously, agents of different architecture do not play against each other.
To reduce the variance and safeguard against (un)lucky runs, we run this procedure $5$ times and average the results over the total of $30$ agents.
Figure~\ref{fig:generalise_all} shows the generalisation reward curves.
The difference between different agents is now stark, with some failing to generalise almost completely.
Both baselines perform barely above chance (zero).
OPRE has the best generalisation, followed by OPRE mix PG with half of its score.

\begin{figure}[h]
\includegraphics[width=\columnwidth]{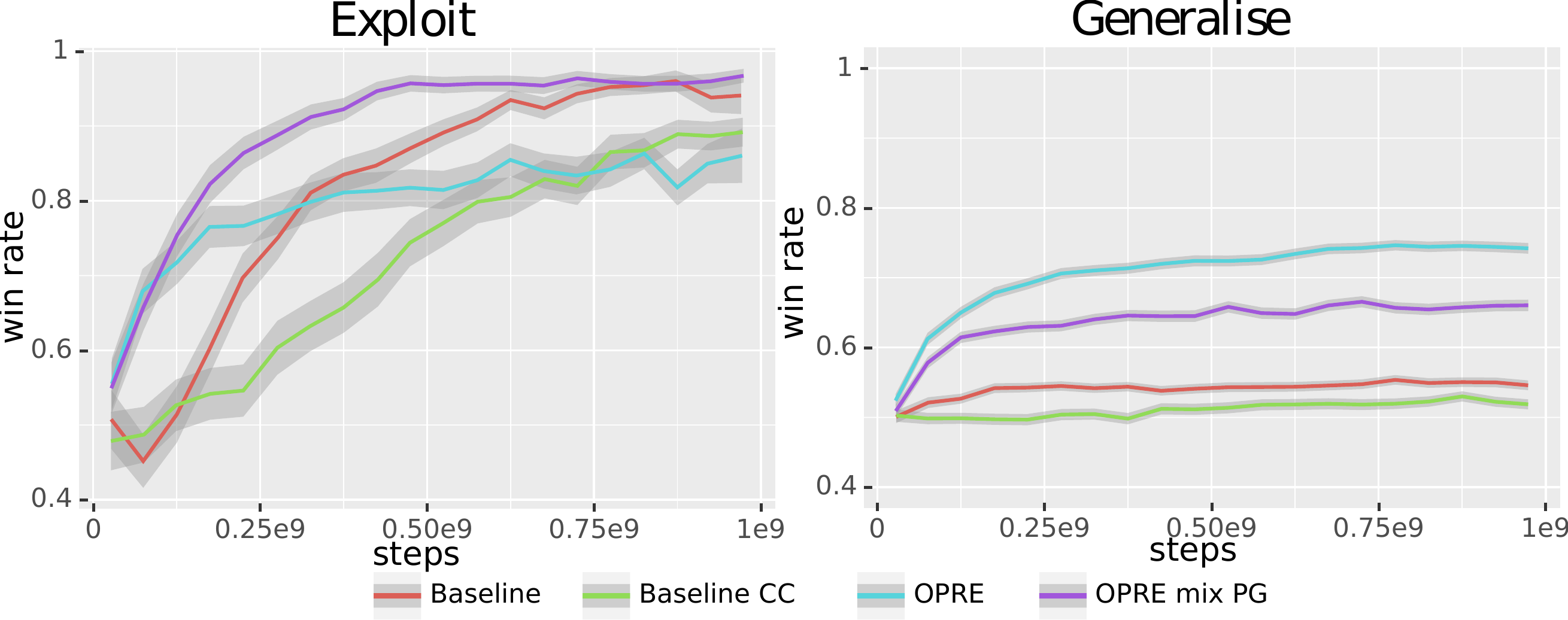}
\caption{ Probability of winning on \textbf{RWS} at a training step. \textbf{Left:} exploiting held-out set; \textbf{Right:} generalising to the held-out set from self-play. \label{fig:RWS_vic_all} \figshrinky}
\end{figure}

\begin{figure}[h]
\includegraphics[width=\columnwidth]{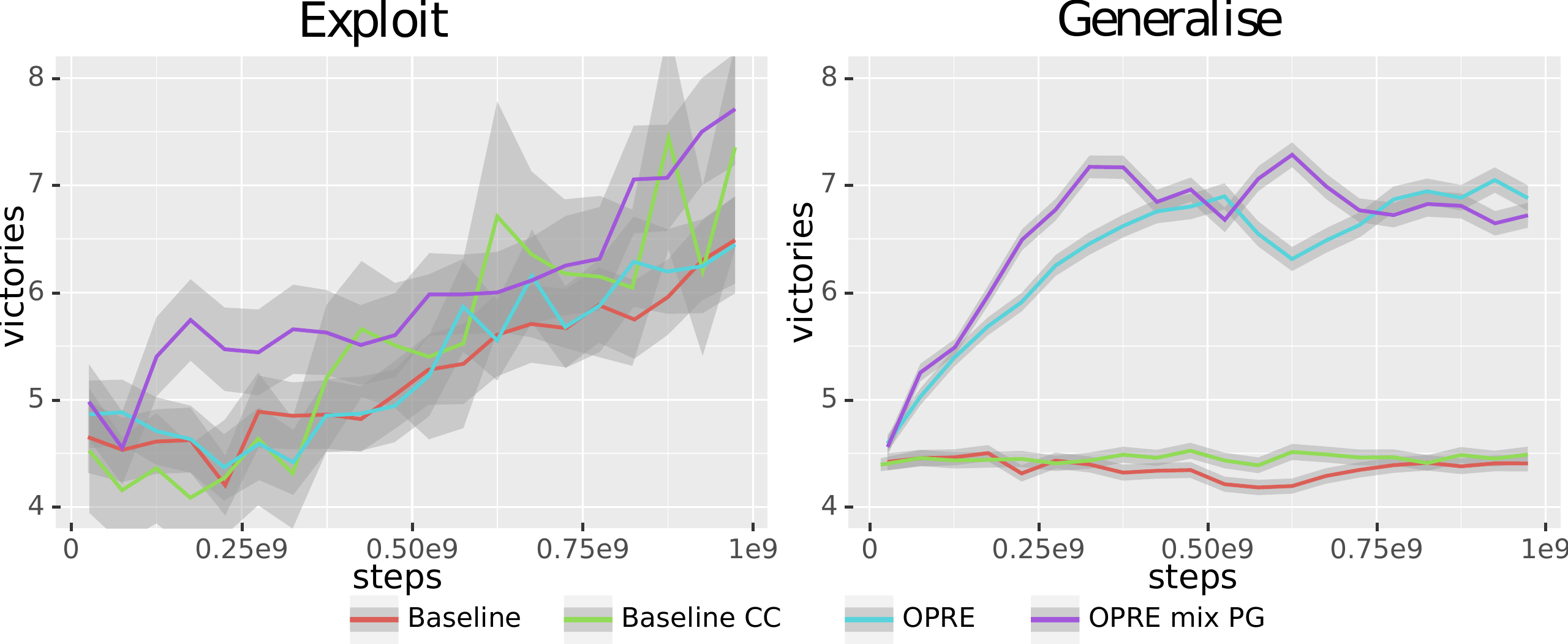}
  \caption{ Reporting average number of victories per episode of \textbf{RPS Arena} at a training step. \textbf{Left:} Exploiting the held-out set; \textbf{Right:} Generalising to the held-out set from self-play. \label{fig:RWA_vic_all} \figshrinky}
\end{figure}

Notice, however, that even OPRE recovers only a part of its return magnitude, compared to exploiting the hold-out directly (Figure~\ref{fig:exploit_all}).
To investigate this gap further we present win rates, instead of return for both the exploitation and the generalisation cases.
Figure~\ref{fig:RWS_vic_all} shows the probability of a victory for an agent in RWS (episodes always end after a confrontation), while Figure~\ref{fig:RWA_vic_all} presents an average amount of victories in a confrontation for RPS Arena (an episode can have multiple confrontations).
OPRE recovers most of its victories even when generalising from self-play, while none of the baselines manage to move significantly above the performance of a random agent.
What gives rise to the difference between return and victory rate is the value of winning a confrontation. 
To maximise the return against a pure strategy, say rock, the agent has to collect as much paper as possible and not a single other resource.
This is the strategy that is learnt by agents trained to exploit the hold-out directly. 
OPREs trained in self-play are more conservative, winning confrontations by a small margin.
They do understand the fundamental, game-theoretic structure of the game, which allows them to win fights, but pursue an overall more conservative policy, since in self-play they face a much stronger adversary - themselves.
This confirms that OPRE successfully extracts the underlying non-transitive reward structure and uses it for generalisation.

\subsection{Ablations}
\label{subsec:exp:ablations}

\begin{figure}[h]
  \centering
{\includegraphics[width=\columnwidth]{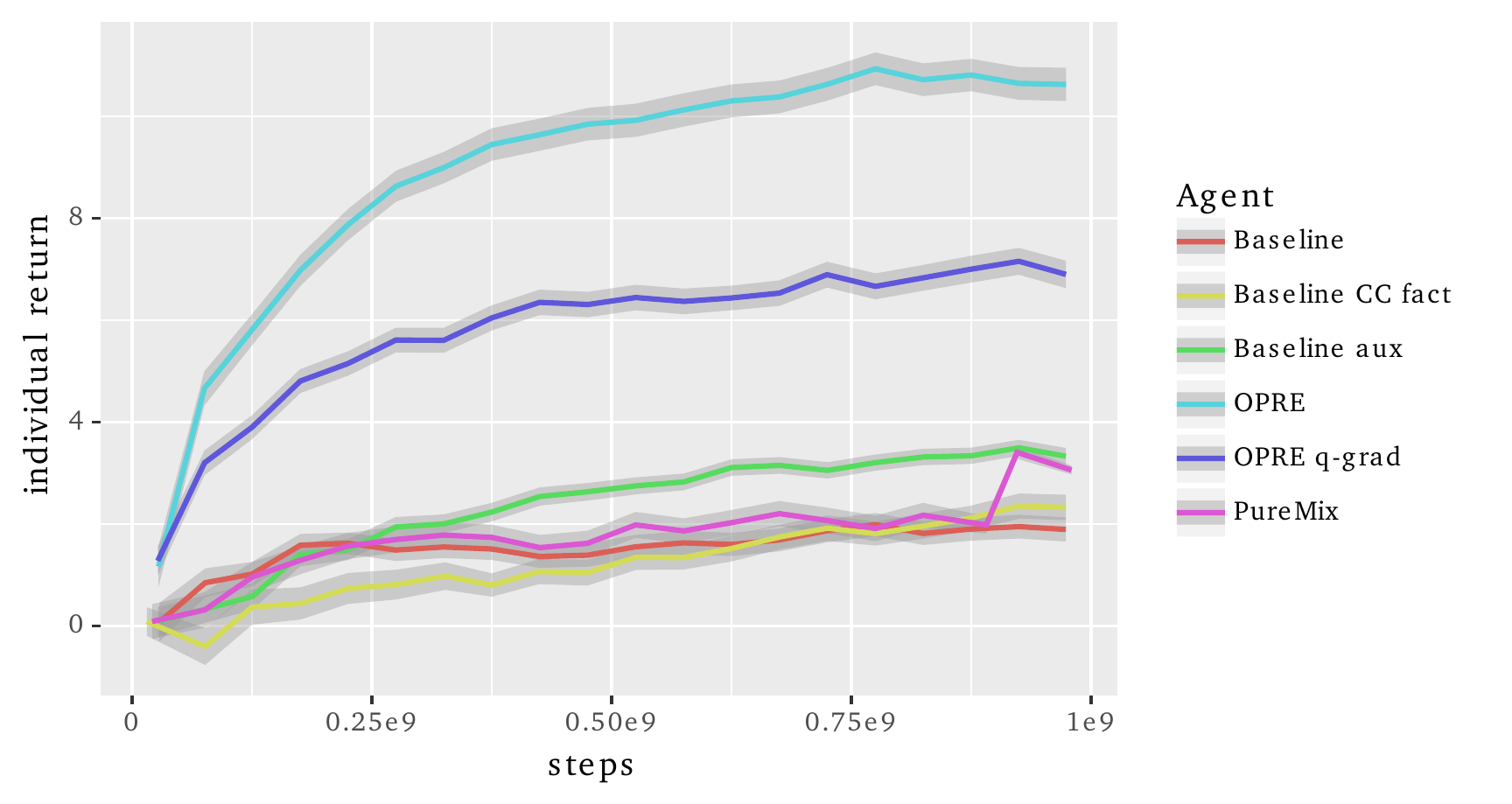}}
  \caption{ OPRE and ablations on \textbf{RWS}. Generalisation to hold-out set from self-play. \figshrinky \label{fig:ablations}}
\end{figure}

In this section we evaluate several ablations of OPRE and baselines, to assess where the improvements come from. 
We focus on the generalisation to the hold-out set of agents trained in self-play on RWS.
Each experiment answers a particular question and corresponds to a curve in Figure~\ref{fig:ablations}.
First, we allow the gradients from $\pi$ to flow into $q(z)$ to check if isolating $q(z)$ from policy gradient and learning it purely from the value function and KL link to $p(z)$ facilitates credit assignment (\textbf{OPRE q-grad}).
To check if the benefit comes from using a mixture model for the policy, we introduce \textbf{PureMix}, which has the same factorised policy structure as OPRE, but uses standard policy gradient to learn it. 
Next, we verify if adding the auxiliary task~\cite{jaderberg2016unreal} of predicting opponents' inventories would be sufficient for the baseline to match OPRE's performance (\textbf{Baseline aux}).
We check if having a mixture value function alone, without a hierarchical policy, is enough to achieve good generalisation. 
To this end we introduce \textbf{Baseline CC fact}, which has the same factorised value function in the critic as OPRE. 
We also estimate $p(z)$ from the agent's own observation and use $KL(q||p)$ as a regulariser, to stay as close to OPRE as possible.
As Figure~\ref{fig:ablations} demonstrates, all the ablations fall significantly short of the full model.

\sasha{Add pure mixture and distrall ablations}

\subsection{Direct agent tournament}
\label{subsec:exp:tournament}

\begin{figure}[h]
\begin{subfigure}{.5\columnwidth}
  \includegraphics[width=1\columnwidth]{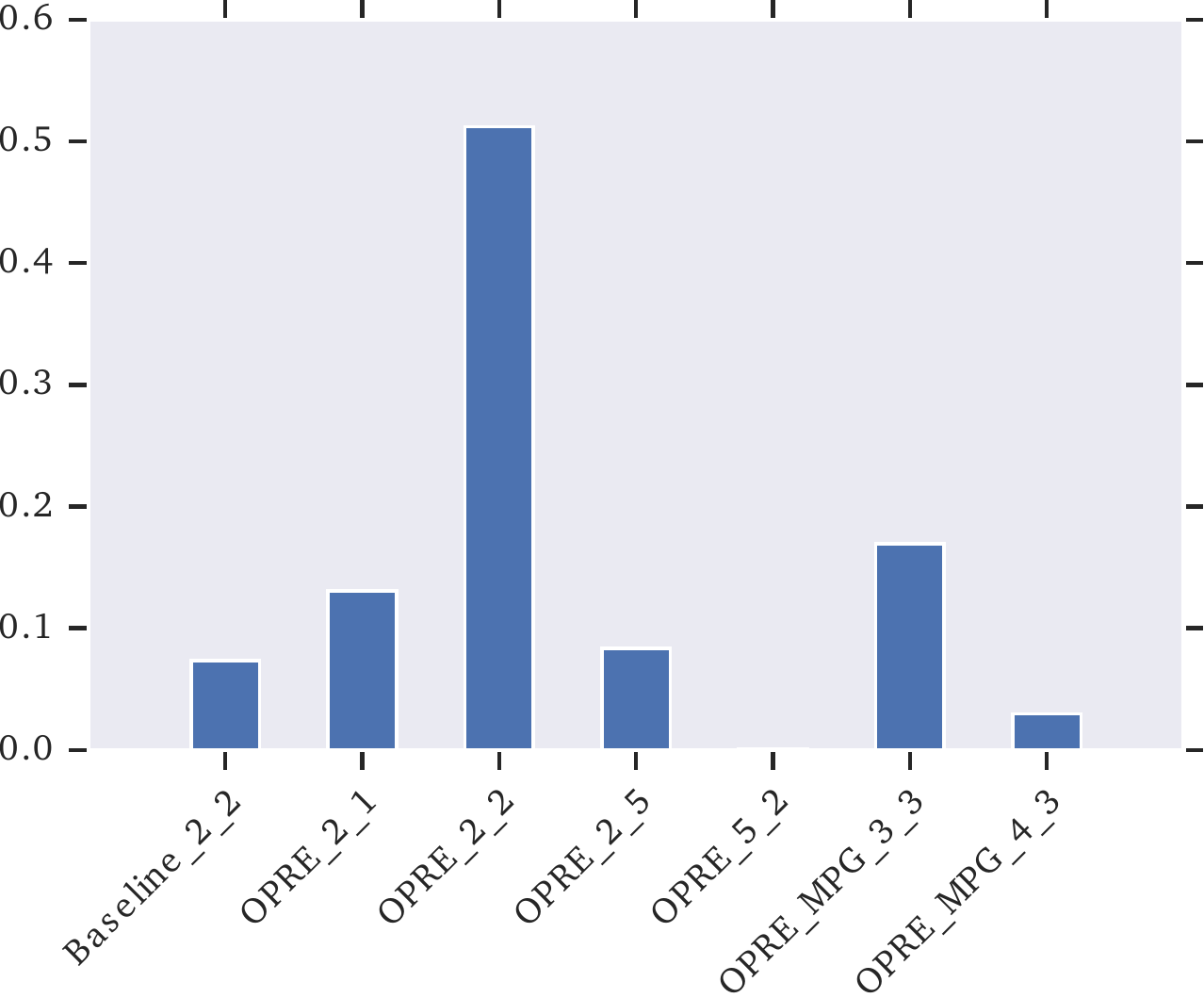}
  \caption{Agents weights in the Nash}
  \label{subfig:nash}
\end{subfigure}%
\begin{subfigure}{.5\columnwidth}
  \includegraphics[scale=0.272]{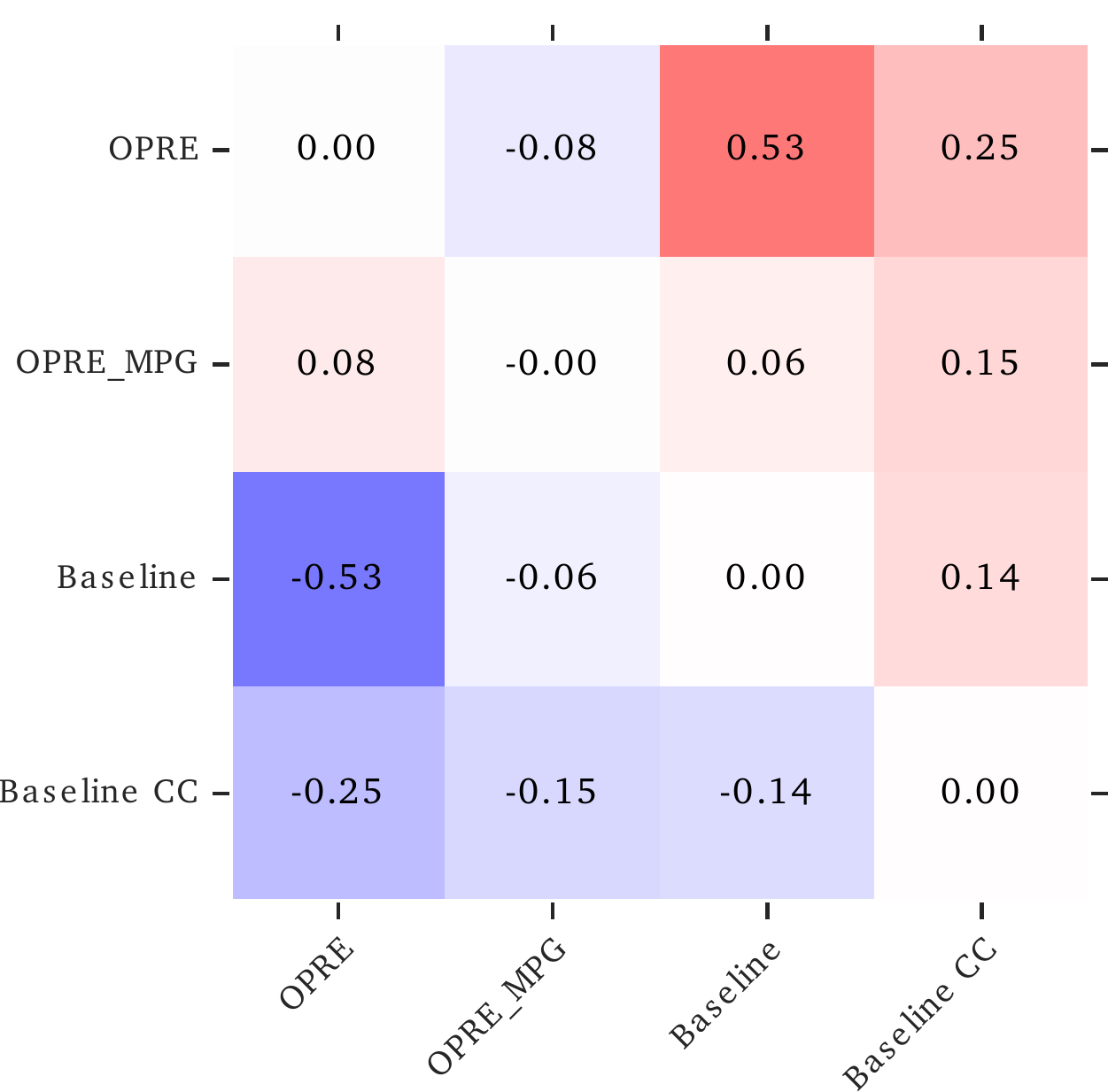}
  \caption{Pay-off between agents}
  \label{subfig:payoffm}
\end{subfigure}
  \caption{ Analysis of direct agent tournament on \textbf{RWS}. \label{fig:crossplay} \figshrinky}
\end{figure}

The zero-sum, two-player setting of RWS allows us to compare different agents in the most direct way -- by playing them against each other.
The experiments in Section~\ref{subsec:exp:generalise} have generated $120$ individual agents ($30$ per agent architecture).
In this Section we are going to look at the pay-off matrix (see appendix for full matrix) generated by playing all of those agents against each other and held-out pure strategies (about 20 million games in total).
The top 16 agents are OPRE (including 3 OPRE mixed PG). 
In Figure~\ref{subfig:payoffm} we plot a pay-off matrix between architectures---a mean pay-off when playing a random agent of one architecture against another, and again OPRE comes out on top.
This confirms that OPRE generalises well not only to a hold-out set of pure strategies but also to other architectures trained in self-play.

\begin{figure}[h]
  \centering
{\includegraphics[width=\columnwidth]{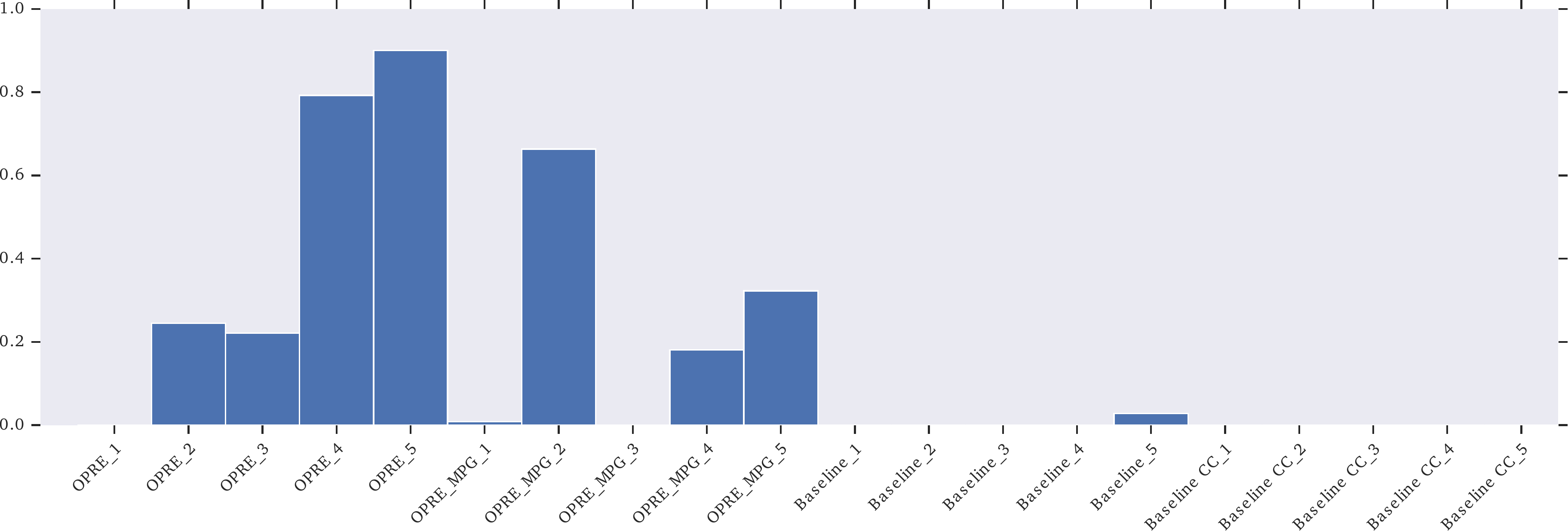}}
  \caption{ Effective diversity of agent populations on \textbf{RWS}. Notice that OPRE agents have high effective diversity compared to baselines. \figshrinky \label{fig:ED}}
\end{figure}

Now we take a deeper look at the pay-off matrix.
Consider a meta-game, defined by the pay-off matrix. 
Figure~\ref{subfig:nash} shows agents weights in the Nash equilibrium of the meta-game. 
Again, all but one of agents in the Nash are OPRE and OPRE mixed PG.
We would like to note here, that Nash equilibria are not unique thereby membership in one should not be used as proof of success, but it is still a positive sign.
Notice that the pay-off matrix has a lot of structure in the OPRE vs OPRE block, i.e. there is no single agent that dominates the population as most agents win against some and lose against the others.
We verify this observation quantitatively by measuring the \textit{effective diversity} (Figure~\ref{fig:ED}), proposed in~\cite{balduzzi2019open}.
Effective diversity quantifies how the best agents (those with support in the Nash) exploit each other. 
If there is a dominant agent then it is zero.
This is the evidence towards OPRE producing more diverse populations than standard architectures, and regular OPRE more diverse than OPRE mix PG. 
This could be another reason for better generalisation, as agents get to experience a larger part of the strategy space in training.

\subsection{Neural analyses}
\label{subsec:neuralrep}

\begin{figure}[h]
  \centering
  \includegraphics[width=\columnwidth]{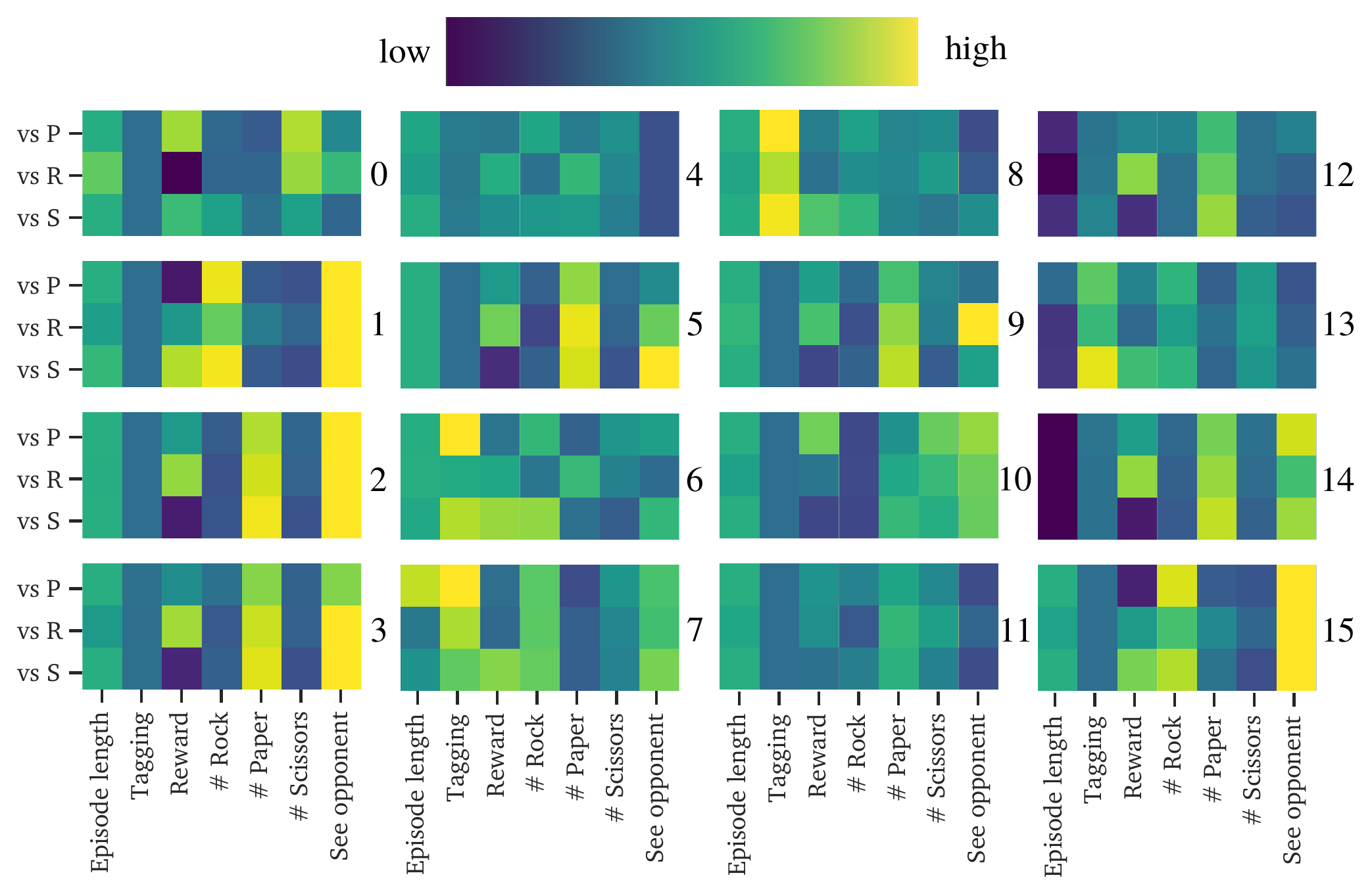}
  \caption{{Analysis of options learnt by OPRE in self-play. 
  Each plate represents an option and each row corresponds to an opposing pure strategy and each column to a behavioural measurement. 
  The colour of the element represents the magnitude of the corresponding measurement (brighter=higher).
  Notice that options 0 focuses on collecting scissors, 1 on rock and 2 on paper.
  Some options include strong scouting component: 1,2,15.
  Option 4 induces tagging.
  Several options have combined effects (2 scouts and collects paper) or duplicate each other (2 and 3).
  \remi{TODO: We need to add some sort of legend here, the colours are not enough I feel. Is yellow higher or lower than green (is brightness or a color scheme indicating the scale?). Also, explicitly naming a couple of options would help.}}}
  \label{fig:Heatmaps}
\end{figure}

To test whether options are diverse low-level policies, we analyse the behaviour of each individual option and verify that it represents some meaningful low-level behaviour, whereby options differ significantly from each other.
To assess which behaviours are encoded by each option, we replace the high-level policy $p$ by a 1-hot vector that places all weight on a single option. 
This transfers behavioural control to a single option for the whole episode. 
We use pure strategies from hold-out set as opponents.
We repeat the process for each option, rolling out 100 episodes per option for each opponent (rock, paper, scissors) and record the following behavioural data: i) episode length ii) tagging events iii) reward iv) collected resources v) scouting---amount of steps where OPRE was directly observing the opponent.
We perform this analysis on an OPRE agent trained in self-play. 
This has an analogue in experimental neuroscience where we activate specific neurons in the living organism to assess which behaviours these neurons encode.
Figure~\ref{fig:Heatmaps} demonstrates that indeed, OPRE's options divide the strategy space into meaningful components, including exploratory ones (e.g., watch opponent), and exploitative ones (e.g., find and collect paper, find and collect rock, find and tag opponent).
Specific behaviours can easily be elicited by activating the corresponding option, and different options often encode opposite behaviours (e.g., tag versus avoid tagging; collect rock versus collect paper).
Please see the appendix D for more detailed analysis of scouting behaviour.

\section{Conclusions}
\label{sec:conclusions}
\secshrinky

In this paper we investigated competitive multi-agent games with concealed information and non-transitive reward function.
We introduced OPRE -- a new agent, which factorises the value function over the latent representation of other agents' concealed information, and then reuses this latent representation to factor the policy into options. 
It only requires access to concealed information for learning the value function during training and never uses it to generate behaviour.
Our experiments on two novel environments demonstrated that OPRE has excellent generalisation from self-play to a hold-out set of unseen opponents and in direct agent tournament.
The analysis of learnt options show that they are meaningful and cover the space of strategies well.

Key insights of this paper can be, potentially, applied to any domain where concealed information is available in hindsight.
As many domains fit this description, this may have implications beyond multi-agent environments.
One exciting direction for future work is extending OPRE to multi-task learning with concealed task definition, where the agent has to both figure out which task it is solving as well as how to solve it.
Another promising idea consists in performing the factorisation not only on the value function, but also on the environment dynamics through, for example, generalised value functions~\cite{sutton2011horde}.

\section*{Acknowledgements}

Authors would like to thank Simon Osindero for inspiring conversations and feedback on the manuscript and John Agapiou for help with experimental setup.

\medskip

\small

\bibliography{deeprl}
\bibliographystyle{icml2020}

\end{document}